\documentclass[mlabstract,onecolumn]{jmlr}
\usepackage{booktabs}
\usepackage{float}     
\graphicspath{{figures/}}

\jmlrproceedings{NeurIPS}{Accepted at NeurIPS}
\jmlrpages{}
\jmlrvolume{}
\firstpageno{1}
\jmlryear{2026}
\jmlrworkshop{Workshop on Symmetry and Geometry in Neural Representations (NeurReps), non-archival}

\newcommand{\A}{\mathcal{A}}
\newcommand{\fmap}{f}
\newcommand{\zt}{z^{t}}\newcommand{\ztp}{z^{t+1}}\newcommand{\zstar}{z^{\star}}
\newcommand{\qt}{q^{t}}
\newcommand{\rhoJ}{\rho(J)}\newcommand{\sigJ}{\sigma_{\max}(J)}
\newcommand{\tdec}{t_{\mathrm{dec}}}\newcommand{\gradz}{\nabla_{z}H}

\title[A fixed-point curve in the CTM]{Riding the Fixed-Point Curve: \titlebreak
Image Classification in the Continuous Thought Machine}

\author{\Name{Thomas Hannagan} \Email{thomas.hannagan@stellantis.com}\\
  \addr Stellantis}

\begin{document}
\maketitle

\begin{abstract}
The Continuous Thought Machine (CTM) is a recurrent network that attends to inputs and makes predictions based on the synchronization between its neurons. On tasks spanning many domains, it displays emergent thought processes that have not been explained. We report a number of findings about how the CTM classifies images, focusing on ImageNet. Thinking steps are attracting but strongly non-normal maps exhibiting balanced amplification (Murphy \& Miller, 2009); while attraction is already present at CTM initialization, training highly promotes non-normality. By Banach's theorem, the neural state thus has a locally unique attracting fixed point, which moves at every step, and the thought trajectory is the curve these fixed points trace, followed one tick behind. Most of that curve's motion is shared across images, and visible as a previously reported traveling wave of activity across neurons. The attention schedule along the thought is largely generic: shuffling queries across images at every tick costs only 7\% of classification accuracy, while image-independent increments predict the next query with $R^{2}=0.98$ (held out). Our results suggest that the CTM has found a single strategy to process all ImageNet images.
\end{abstract}

\section{Introduction}
The Continuous Thought Machine \citep{darlow2025continuous} evolves a neural state for a fixed number of
internal ticks against frozen image features, choosing at each tick where to attend and reading a
prediction from the \emph{synchronization} between neurons --- a decayed second moment of their
co-activations. It has been applied to classification, mazes, parity, sorting and reinforcement learning,
and displays emergent behaviour, including low-frequency traveling waves across its neurons, that has not
been explained. We take the official ImageNet checkpoint, without further training, and ask what kind of
object the internal trajectory is.

The CTM's neural synchronization latent acts like a memory: when it is held fixed, a tick is an ordinary map whose local behaviour is
read from one Jacobian; that map is almost always contractive in spectrum but strongly non-normal, so it has a locally unique
attracting fixed point that the state never quite reaches. Because the memory changes each tick, the
target moves, and inference is a curve of moving fixed points followed one step behind
(\S\ref{sec:fp}). Most of the curve's motion is shared across images, and attention follows a schedule that is shared as well (\S\ref{sec:carrier-main}). Ordering neurons by the phase of that curve reproduces a traveling wave of neural activity reported by the CTM authors
(\S\ref{sec:waves}), and the image's class rides on the curve from the first tick (\S\ref{sec:class}). We then turn to the memory's role in sharpening the prediction (\S\ref{sec:entropy}) and to the contrast with smaller MNIST and CIFAR models (\S\ref{sec:small}).

\section{The Continuous Thought Machine}\label{sec:setup}\label{sec:bg}\label{app:ctm}
We use the official ImageNet CTM ($D{=}4096$ neurons, $T{=}50$ ticks, a ResNet backbone, $L{=}196$
attention tokens); it reaches $0.72$ top-1 on our validation subsample (one image from each of the $1000$ ImageNet classes of the validation set). All ImageNet results in this paper are on this subsample or subsets of it, except for class identity (\S\ref{sec:class}), which uses its own sample of $50$ images per class; sample sizes and 95\% intervals for every headline number are in Appendix~\ref{app:controls}.

\begin{figure}[H]
\centering
\includegraphics[width=0.85\linewidth]{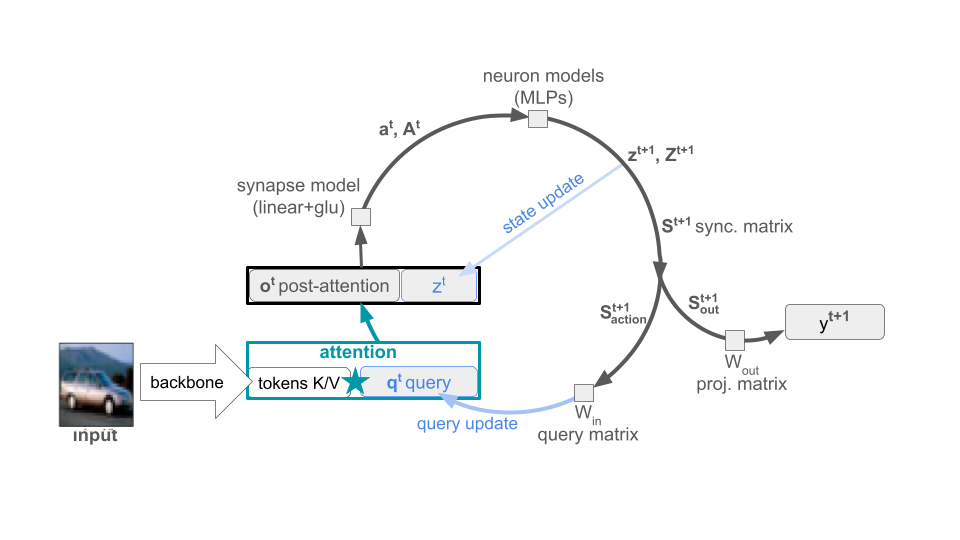}
\caption{\textbf{One tick of the CTM.} After a single backbone pass the model iterates against a frozen
token bank: a query $\qt$ read from the action-side synchronization slice drives attention; a synapse
model mixes the attention output $o^{t}$ with the state $\zt$; per-neuron networks produce the next
state $\ztp$; the synchronization matrix $S^{t+1}$ is updated and its output slice is read to logits
$y^{t+1}$.}
\label{fig:arch}
\end{figure}

We follow the notation of \cite{darlow2025continuous}, generally writing matrices in upper-case and vectors in lower-case, with tick indices as superscripts. As illustrated in Fig.~\ref{fig:arch}, the image is first passed
through a convolutional backbone network, which extracts features into a key/value bank K:
\begin{equation}
K \;=\; \mathrm{proj}\big(\mathrm{flat}(b_\phi(x))\big) \;\in\; \mathbb{R}^{L\times d},
\label{eq:bank}
\end{equation}
this is a matrix of $L$ tokens of width $d$. K is computed once and held fixed: the model never re-observes the input. It then runs $T$ internal ticks against $K$, carrying a vector of post-activations $z^{t} \in \mathbb{R}^{D}$ --- $D$ neurons in total --- from a learned, input-independent initial value $z^{1}$.
One tick proceeds in four steps.

\paragraph{Attending.} The current post-activations are summarized by the action-side latent representation $S^{t}_{\mathrm{action}}$ (selected entries of $S^{t}$), which a linear map $W_{\mathrm{in}}$ turns into a query $q^{t}$ that modulates the data by cross-attention:
\begin{equation}
q^{t} = W_{\mathrm{in}}\,S^{t}_{\mathrm{action}} \;\in\; \mathbb{R}^{d},
\qquad
o^{t} \;=\; \mathrm{softmax}\!\left(\frac{q^{t} K^{\!\top}}{\sqrt{d}}\right) K \;\in\; \mathbb{R}^{d}.
\label{eq:attn}
\end{equation}
This is the sole coupling between the thought and the image: $K$ is fixed, and the query depends on the trajectory only through $S^{t}_{\mathrm{action}}$.

\paragraph{Updating the state.} The attention output is concatenated with the current post-activations and passed through the \emph{synapse model} $f_{\theta_{\mathrm{syn}}}$, a single
gated linear layer shared across neurons, giving the \emph{pre-activations} $a^{t} \in \mathbb{R}^{D}$; this
is the only step in which neurons exchange information. The \emph{neuron-level models}
$g_{\theta_d}$ map pre-activations to the next post-activations, each neuron seeing only its own history
$A^{t}_{(i)} \in \mathbb{R}^{M}$ through its own private MLP. Cross-neuron mixing is thus confined to
$f_{\theta_{\mathrm{syn}}}$, while $g_{\theta_d}$ gives each neuron an individual temporal response --- so the timing of neurons relative to one another becomes a carrier of information, which is what the synchronization matrix later reads:
\begin{equation}
a^{t} = f_{\theta_{\mathrm{syn}}}\big([\,o^{t};\,z^{t}\,]\big),
\qquad
A^{t} = \big[a^{t-M+1},\dots,a^{t}\big] \in \mathbb{R}^{D\times M},
\qquad
z^{t+1} = g_{\theta_d}\big(A^{t}\big).
\label{eq:step}
\end{equation}

\paragraph{Accumulating the memory.} The neural states $z^{1},\dots,z^{t}$, stacked into $Z^{t} \in \mathbb{R}^{t\times D}$ are used to build the \emph{synchronization matrix}: it is the running second moment between neurons,
\begin{equation}
S^{t} \;=\; \tfrac{1}{t} \big(Z^{t}\big)^{\!\top} Z^{t} \;=\; \tfrac{1}{t}\sum_{\tau \le t} z^{\tau} \big(z^{\tau}\big)^{\!\top}
\;\in\; \mathbb{R}^{D\times D}.
\label{eq:st}
\end{equation}

\paragraph{Reading a decision from it.} The model does not store $Z^{t}$: it keeps $S^{t}$ as a running accumulator, with an
exponential decay $r_{ij} = \exp(-\delta_{ij})$ learned per neuron pair --- one scalar each, fixed after training and identical for every input and every tick.
From the synchronization matrix $S^{t}$, two latent representations are extracted, $S^{t}_{\mathrm{out}}$ and $S^{t}_{\mathrm{action}}$, which are simply fixed slices of its entries and respectively determine the prediction (out) and the next features to attend to (action). The out-side latent representation $S^{t}_{\mathrm{out}}$ is read out by a linear map $W_{\mathrm{out}}$ to that tick's logits $y^{t}$ and certainty $\mathcal{C}^{t}$, and the action-side $S^{t}_{\mathrm{action}}$ is projected through a linear layer to form the next query:
\begin{equation}
y^{t} = W_{\mathrm{out}}\,S^{t}_{\mathrm{out}},
\qquad
\mathcal{C}^{t} = 1 - H(y^{t}),
\qquad
q^{t+1} = W_{\mathrm{in}}\,S^{t}_{\mathrm{action}}.
\label{eq:out}
\end{equation}

At $t=1$ there is no history to summarize: $z^{1}$ and the initial pre-activation trace are learned parameters identical for every image, so the first attention read is the model's learned prior over where to look. In other words, every thought trajectory starts from the same point.

\paragraph{Training objective.} $T$ is the thought budget: fixed in advance, identical at training and inference, with no halting mechanism and no early exit. The loss is a \emph{dual-tick} cross-entropy, averaging the cross-entropy at whichever tick had the lowest loss and at whichever had the highest certainty, both chosen per input. Nothing weights later ticks, requires an answer before $T$, or rewards the synchronization statistic directly. Throughout, each sample's decision tick is its tick of highest certainty, $\tdec=\arg\max_t\mathcal{C}^{t}$, as in \citet{darlow2025continuous}, and every decision-dependent quantity is computed over $[1,\tdec]$. Sample sizes differ between analyses because some (materialized Jacobians, Schur forms) are expensive; the sample size is stated with each result.

\paragraph{Design choices specific to ImageNet.}
Under the official ImageNet checkpoint the synchronization memory uses \emph{random pairing}: rather than forming the full $D\times D$ matrix, a fixed set of neuron pairs $(i,j)$ is drawn once at initialization and only those entries are tracked, so the action and output slices are long vectors rather than a dense matrix. The checkpoint tracks $8196$ pairs for the output slice and $2048$ for the action slice, against $D=4096$ neurons. The sampling is broad rather than selective: a neuron takes part in $4.0$ output pairs on average and $98\%$ of neurons appear in at least one, so the readout draws on the population as a whole rather than on a privileged subset. None of this affects any state-space quantity used here.

\paragraph{What is known about how the CTM forms decisions.}
\label{sec:known}
The original CTM paper reports several findings on ImageNet-1K~\citep{darlow2025continuous} at a budget of
50 ticks. Most images pass a $0.8$ \emph{certainty} threshold within ten ticks with a long tail out to
fifty, so the budget could in principle be halted early, as in adaptive-computation-time recurrent
networks~\citep{graves2016act}; the logits are unusually well calibrated without post-hoc adjustment; and
reading the prediction at the most certain tick beats reading the current one beyond roughly fifteen
ticks, while averaging logits does worst---so the model passes through low-certainty, sometimes incorrect
states before settling. That certainty rises---equivalently, that output entropy falls---along the thought
in most cases is therefore already known, and is in fact built into the CTM's objective. The attention
trajectory over image features also emerges through training, and visual inspection shows diverse,
multiscale dynamics with low-frequency traveling waves across a learned feature map arising with no
driving signal---a phenomenon with a long history in cortex~\citep{muller2018cortical}. One further property of the released ImageNet checkpoint bears on what follows: its learned decay rates are negligible over the $50$-tick budget, so $S_{\text{out}}$ is in effect a running average of the state's co-activations, a deterministic quadratic functional of $z^{1},\dots,z^{t}$. We therefore analyze the state $z^{t}$, where the recurrence's learned per-tick maps and its fixed points live.

\section{An attracting, non-normal fixed-point curve}\label{sec:fp}
What governs the state's motion? One way to study recurrent network dynamics is to locate the
fixed points of the recurrence and linearize there~\citep{sussillo2013opening}. We hold the memory
context $\A^{t}$ fixed, freezing the image tokens, synchronization memory, and pre-activation history at tick $t$, \[ \mathcal A^t \equiv (K,S^t,A^t), \] which turns the update into an ordinary map \[ z^{t+1}=f(z^t;\mathcal A^t). \]

We reconstruct that map exactly (a single reconstructed step reproduces the model's next state to
$<\!10^{-6}$) and iterate it: it converges at $3055$ of $3072$ image--tick pairs from $256$ images (residual $<\!10^{-6}$; App.~\ref{app:fp}), and at every one of these fixed points $\zstar$ the Jacobian $J$, read by autograd power iteration, has spectral radius well below one ($\rhoJ=0.15$
median, 95\% interval $0.14$--$0.16$, max $0.95$). By Banach's theorem, applied in an adapted norm (App.~\ref{app:fp}), the fixed point is locally unique and attracting: wherever the iteration converges, the state has one well-defined target. Beyond the neighbourhood the theorem covers, $3456$ restarts from random states and from other images' states all converged to the same fixed point, and we found no second basin (App.~\ref{app:fp}).

\begin{figure}[H]
\centering
\includegraphics[width=1.0\linewidth]{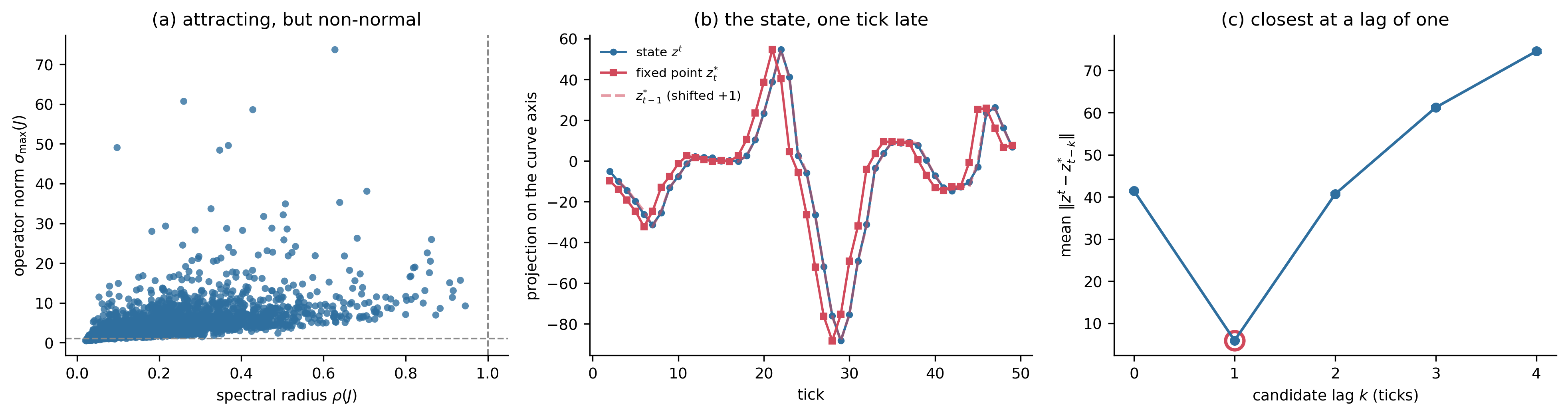}
\caption{\textbf{Attracting but non-normal, and one tick late.} \textbf{(a)}~At each fixed point the spectral radius $\rhoJ$ stays below one (the map is stable and the fixed point unique) while the operator norm $\sigJ$ exceeds one (it is not a Euclidean contraction); the gap is the non-normality. \textbf{(b)}~State and fixed point projected on the leading axis of the fixed-point curve, with $\zstar_{t-1}$ overlaid: the state's trajectory is the fixed-point curve delayed by one tick. \textbf{(c)}~The same claim measured: the mean distance $\|\zt-\zstar_{t-k}\|$ to the fixed point $k$ ticks earlier, against candidate lag $k$, is minimized at $k=1$ ($5.9$, against $41.4$ at $k=0$ and $40.7$ at $k=2$).}
\label{fig:spectrum}
\end{figure}

The map is nevertheless strongly \emph{non-normal}: although every eigenvalue decays, the largest
singular value exceeds one ($\sigJ=3.2$ median, up to $74$), so it briefly stretches a few directions
even while contracting overall (Fig.~\ref{fig:spectrum}a). Estimation details, the per-fixed-point ratio
$\sigJ/\rhoJ$, and a finite-difference confirmation of the stretching are in Appendix~\ref{app:fp}.

\paragraph{What training adds.} Attraction itself comes with the architecture: an untrained CTM also has $\rhoJ<1$ at every fixed point (median $0.05$), as does a \emph{random core}---the trained backbone with the recurrence at initialization, so the recurrence sees the same image features (median $0.24$). Against this matched-input control, training lowers the spectral radius ($0.24\to0.15$) and raises the non-normality $\sigJ/\rhoJ$ from about $7$ to about $23$, while the untrained map is a plain Euclidean contraction ($\sigJ<1$ everywhere). Training also concentrates each neuron model's sensitivity on the newest entry of its pre-activation window, where $\zt$ enters ($2.6\times$ its uniform share, against uniform at initialization), so the contraction is not an artifact of $\zt$ entering only one of $M$ window slots.

\begin{figure}[H]
\centering
\includegraphics[width=1.0\linewidth]{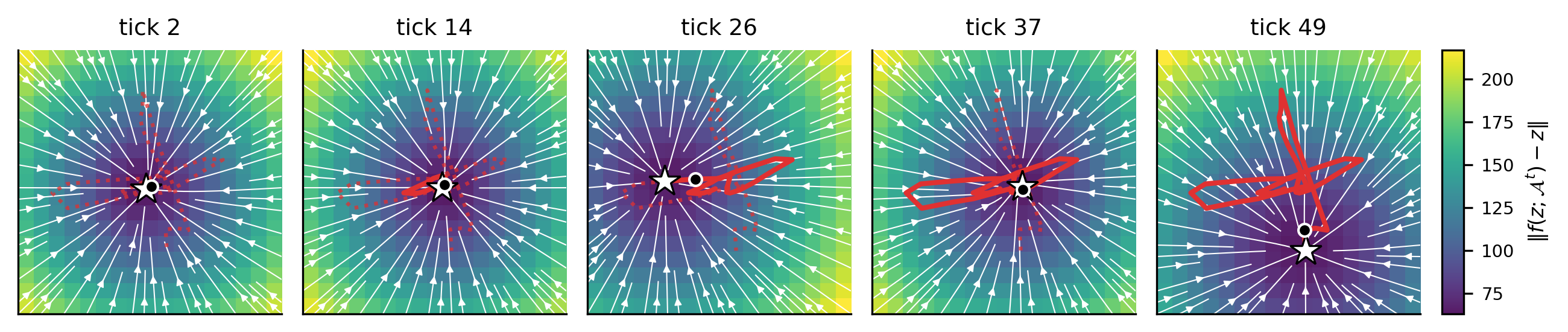}
\caption{\textbf{The fixed-point curve.} The recurrence's residual field on the plane spanned by the
path of successive fixed points, shown across ticks for one held-out image. Color is residual magnitude,
streamlines the in-plane flow; the red curve is the path of the fixed point $\zstar$, the star its
current location, and the black dot the state, which stays within about one step of the moving
attractor.}
\label{fig:fpcurve}
\end{figure}

\paragraph{One tick behind.} Because the memory changes with every tick, the target moves, and the trajectory is simply the curve
these fixed points trace, followed one update behind (Fig.~\ref{fig:spectrum}b). The state is not on the current fixed point but on the previous one: the mean distance $\|\zt-\zstar_{t-k}\|$ is minimized at $k=1$ and is then seven times smaller than the distance to the current fixed point ($5.9$ versus $41.4$, over $128$ images; Fig.~\ref{fig:spectrum}c). The offset points backwards along the fixed point's own motion (median $\cos(\zt-\zstar_{t},\,\zstar_{t-1}-\zstar_{t})=0.995$) and measures a full step of it (median ratio $1.01$), so the state trails the moving attractor by one update rather than sitting on it (Fig.~\ref{fig:fpcurve} shows the trailing state). This lag is largely structural: $\zt$ is one contraction step from $\zstar_{t-1}$, and the shrink factor across one step (median $0.11$ over $1536$ image--tick pairs) is what $\rhoJ$ at the same fixed points predicts (median $0.14$); what is not structural is how far the fixed point moves each tick, which keeps the state a full step behind (App.~\ref{app:fp}).

\paragraph{The non-normality is balanced amplification.} Schur-triangularizing
$J=UTU^{*}$, almost the whole operator lives in the strictly upper-triangular, feed-forward part: that
fraction --- which equals Henrici's departure from normality --- is $0.99$, against $0.71$ for a Ginibre
matrix and $0.93$ for the random core (equivalently, the eigenvalues carry $2.0\%$ of $\lVert J\rVert_F^2$, against $14\%$ for the random core), while the antisymmetric part sits at the random-matrix level. With $\rhoJ<1$, this triad of
near-maximal feed-forward non-normality, stable eigenvalues and chance-level rotation is the structure
\citet{murphy2009balanced} identified as \emph{balanced amplification}: a hidden feed-forward chain that
amplifies activity in transit while every mode decays (App.~\ref{app:schur}).

\section{Attention follows a shared schedule}\label{sec:carrier-main}
\begin{figure}[H]
\centering
\includegraphics[width=1.0\linewidth]{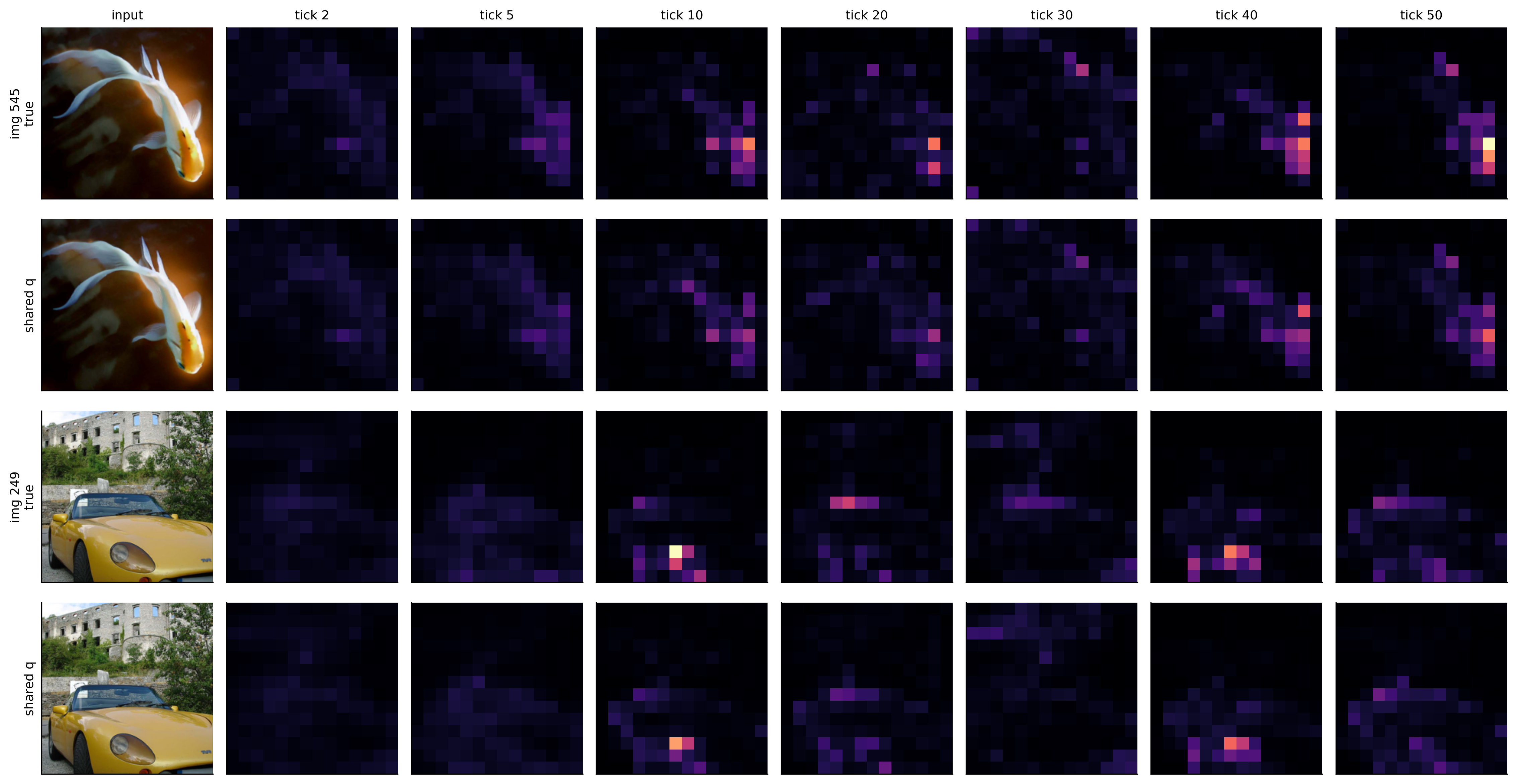}
\caption{\textbf{The attention scan is a shared schedule.} For two held-out images (input, first column),
the model's true attention (first row of each pair) and the attention produced by the \emph{shared} population-average
query (second row), across ticks; one color scale per image. The shared schedule recovers the image-specific scan.}
\label{fig:scan}
\end{figure}

Where the model attends is a program mostly shared across images, rather than an image-specific search. Attention is selective: it concentrates on about $96$ of the $196$ tokens (as measured by the exponential of its entropy), with its most-attended token carrying $11\times$ the uniform weight. \textbf{Ablations} (Table~\ref{tab:ablation}): freezing the query at
its first-tick value, so the model never advances its schedule, removes $34\%$ of accuracy, and
flattening attention removes most of it ($-80\%$); handing each image \emph{another image's}
query at the matching tick costs only $7\%$ ($1000$ images). The progression is therefore essential, but its image-specific
content matters far less. \textbf{Factorization:} substituting the population-average query at each tick reproduces
the image's own attention map at cosine $0.936$ (Fig.~\ref{fig:scan}) --- a generic schedule pointed at
the image's features. \textbf{Prediction:} the current query plus the shared average increment forecasts
the next attention map at cosine $0.996$, against $0.948$ for assuming attention stays put ($400$ held-out images): the next
glance is predictable before the model takes it (App.~\ref{app:carrier}). This is a one-step-ahead forecast: the predictor consumes the current query, so it anticipates the next glance from the current one, not the whole scan from the image alone.

\begin{table}[H]
\centering\small
\caption{Query/attention ablations on ImageNet (top-1 at $\tdec$; chance $0.001$). Freezing the query's
schedule is costly; replacing its image-specific content costs little. $n{=}1000$ images; 95\% intervals in Appendix~\ref{app:controls}. Without attention the image never enters after the backbone and $z^{1}$ is a learned constant, so every image follows the same trajectory and receives the same class; accuracy is that class's frequency in the sample.}
\label{tab:ablation}
\begin{tabular}{lccccc}
\toprule
 & intact & frozen ($q^{1}$) & shuffled (cross-image) & uniform & no-attention \\
\midrule
top-1        & $0.723$ & $0.480$ & $0.675$ & $0.145$ & $0.001$ \\
rel.\ change & --- & $-34\%$ & $-7\%$ & $-80\%$ & $-100\%$ \\
\bottomrule
\end{tabular}
\end{table}

\textbf{What the schedule does to the curve.} Most of the fixed point's motion is common to all images: the mean curve over images carries $77\%$ of its squared motion per tick. Rolling the model out with an imposed query and recomputing the fixed points ($96$ images), freezing the query at its first-tick value leaves $97\%$ of the fixed point's motion, both overall and in each image's own part. Handing each image another image's schedule moves its curve about a third of the way towards that image's curve ($0.35$ of the distance between the two), while each step keeps the direction of the image's own (cosine $0.86$ once the shared curve is removed, against $0.00$ with the other image's step). The schedule thus shifts each image's curve, which matters for the decision, while the curve's motion comes mostly from the recurrence and its memory.

\section{The fixed point curve is a traveling wave}\label{sec:waves}\label{app:waves}
\begin{figure}[htbp]
\centering
\includegraphics[width=0.85\linewidth]{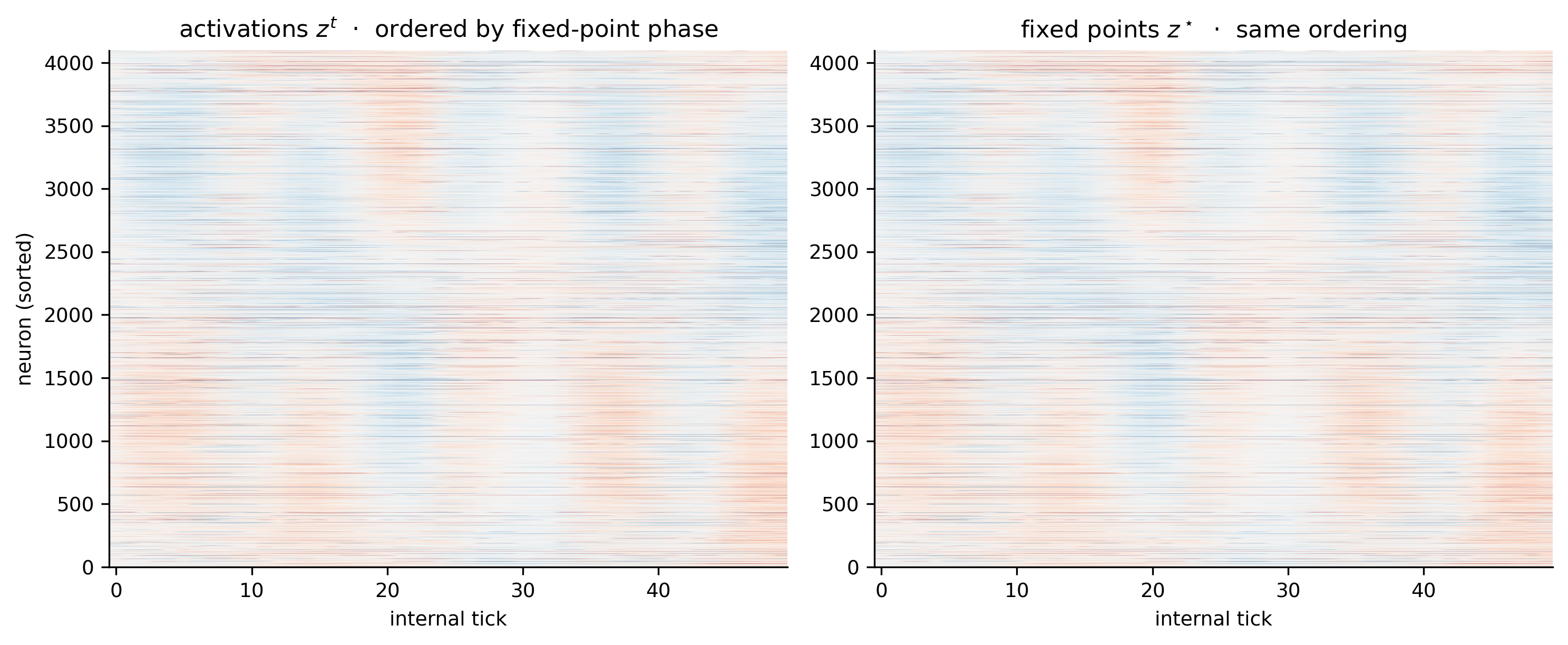}
\caption{\textbf{The traveling wave is the moving fixed point.} One held-out
image; neurons ordered by the phase of $z^{\star}(A^{t})$. \textbf{Left:} activations $z^{t}$ over ticks
--- the reported wave. \textbf{Right:} the fixed points under the same ordering, indistinguishable from
the left.}
\label{fig:wave}
\end{figure}

The curve (\S\ref{sec:fp}) accounts for a phenomenon the CTM authors report: low-frequency traveling waves
across a UMAP of activation profiles, arising with no driving signal. They are the moving fixed point made
visible. Ordering the $D$ neurons by the phase of the fixed-point trajectory $z^{\star}(A^{t})$ or by their own phase recovers the same traveling wave (Fig.~\ref{fig:wave}, left
vs right). This ordering recovers the authors' activation-phase ordering (peak-tick Spearman $0.82$, circular phase correlation $0.88$; medians over $128$ images) --- as expected, since the state tracks $z^{\star}$ within about one step, so activation phase is essentially fixed-point phase. The neurons' peak activation times are spread across essentially the whole window --- standard deviation $15.0$ of $50$ ticks for both the fixed points and the activations, against ${\approx}14.4$ for a uniform spread --- so neurons traverse the shared motion in staggered phase rather than together, which is what makes the front travel rather than stand. Because $\rhoJ<1$ the fixed point moves smoothly and the resulting pattern is a transient propagating pulse rather than a sustained standing wave, matching the low-frequency, driverless character of the reported phenomenon.

That the two panels match follows largely from the one-step lag of \S\ref{sec:fp}. The wave is not, however, coupled to the attention schedule: across $128$ images, the cross-validated canonical correlation between the fixed point's leading temporal modes and the schedule's ($0.52$) does not exceed a permutation null (95th percentile $0.54$; $p=0.17$). The fixed-point trajectory is also higher-dimensional than a single front: its top-two temporal modes hold only $0.38$ of the variance, on average over these images.

\paragraph{The cascade account, and why it fails.} The balanced-amplification structure of \S\ref{sec:fp} suggests a compelling alternative account of the waves. If
the tick map is an almost-pure feed-forward cascade that also organizes the
observed dynamics, then the traveling waves should appear when neurons are ordered by their position in
that cascade. They do not. Ordering neurons by their position in $J$'s feed-forward chain --- each neuron's mean index in the Schur basis $J=UTU^{*}$, weighted by its participation $\sum_k U_{ik}^{2}$, using $J$ at the mid-trajectory tick --- peak activation time does not advance: the rank correlation between cascade position and peak tick is $-0.08$ over all $D$ neurons, and between cascade position and activation phase $+0.01$ (medians over $16$ images). The reason is structural: $J$ --- and with it its Schur ordering --- is redefined at every tick, so no single instantaneous cascade can organize a trajectory that runs over all ticks; and because the state sits on the moving fixed point, the bulk of the activation is carried by where that fixed point is, while the balanced-amplification transient governs only the one-step residual. The traveling waves are thus a signature of the moving fixed-point curve, not of the instantaneous feed-forward cascade; the balanced-amplification structure of $J$ stands as a property of the operator whose functional role is left open.

\section{The curve carries class identity}\label{sec:class}\label{app:class}
Every trajectory starts from the same learned $z^{1}$, so class identity must enter the curve along the way. Our analyses show that it does so at once.

To determine this, we select $50$ classes with $50$ validation images each ($2500$ images), drawn at random from the $1000$ ImageNet classes. The model is rolled out once per image, recording the state $\zt$, the fixed point $\zstar_{t}$ (iterated to relative tolerance $10^{-5}$) and the readout memory $S^{t}_{\mathrm{out}}$ at $15$ ticks.

\paragraph{When class becomes decodable.} At each tick, a linear probe (standardization, $128$ principal components, multinomial logistic regression; $5$-fold cross-validation) decodes class from each representation. With about $40$ training images per class the probes are a lower bound: the model's own readout, trained on all of ImageNet, reaches $0.92$ where the probes plateau near $0.86$. Comparisons are therefore made between representations under the same probe. The fixed point leads: $0.82\pm0.02$ at tick $1$ ($50$-way; chance $0.02$) against $0.79\pm0.02$ for $\zt$ and $0.79\pm0.03$ for $S^{t}_{\mathrm{out}}$, and above the token-bank mean of the image features ($0.75\pm0.02$). The state's accuracy at tick $t{+}1$ tracks the fixed point's at tick $t$ ($0.81$ vs $0.82$, $0.83$ vs $0.84$, $0.84$ vs $0.85$ over ticks $1$--$4$); this is the one-tick lag of \S\ref{sec:fp} seen in class information. The fixed point and the state peak near $0.86$ by tick $4$--$6$, and $S^{t}_{\mathrm{out}}$ near $0.87$ by tick $10$; thereafter $S^{t}_{\mathrm{out}}$ holds its level while $\zstar_{t}$ and $\zt$ drift slightly lower (to about $0.83$), with transient dips (ticks $25$ and $36$) that we do not interpret. The model's readout, restricted to the $50$ classes, rises from $0.73$ at tick $1$ to $90\%$ of its plateau at tick $3$, as the memory accumulates, and plateaus near $0.92$ from tick $13$.

\paragraph{How class is encoded relative to the curve.} We write $\zstar_{i,t}=c_{t}+d_{k(i),t}+r_{i,t}$, with $c_{t}$ the mean over images (the shared curve), $d_{k,t}$ the offset of class $k$, and $r_{i,t}$ the image residual. Over images and ticks, the shared curve carries $0.66$ of the variance, class offsets $0.105$ (against $0.007$ when labels are shuffled) and the image residual $0.24$. The class offsets lie almost entirely outside the curve's leading two-dimensional plane ($0.5\%$ of their variance inside it, against $0.05\%$ for a random plane): classes do not occupy different regions of the curve but ride the same curve, displaced sideways. The offsets are not fixed in direction either: their cosine with the final offset is near zero at first, about $0.6$ over ticks $6$--$16$, falls to about $0.1$ near tick $25$, and recovers by the end. At the final tick the mean class offset is $21\%$ of the distance travelled along the curve.

\section{What the synchronization memory does}\label{sec:entropy}\label{app:entropy}
\begin{figure}[H]
\centering
\includegraphics[width=1.0\linewidth]{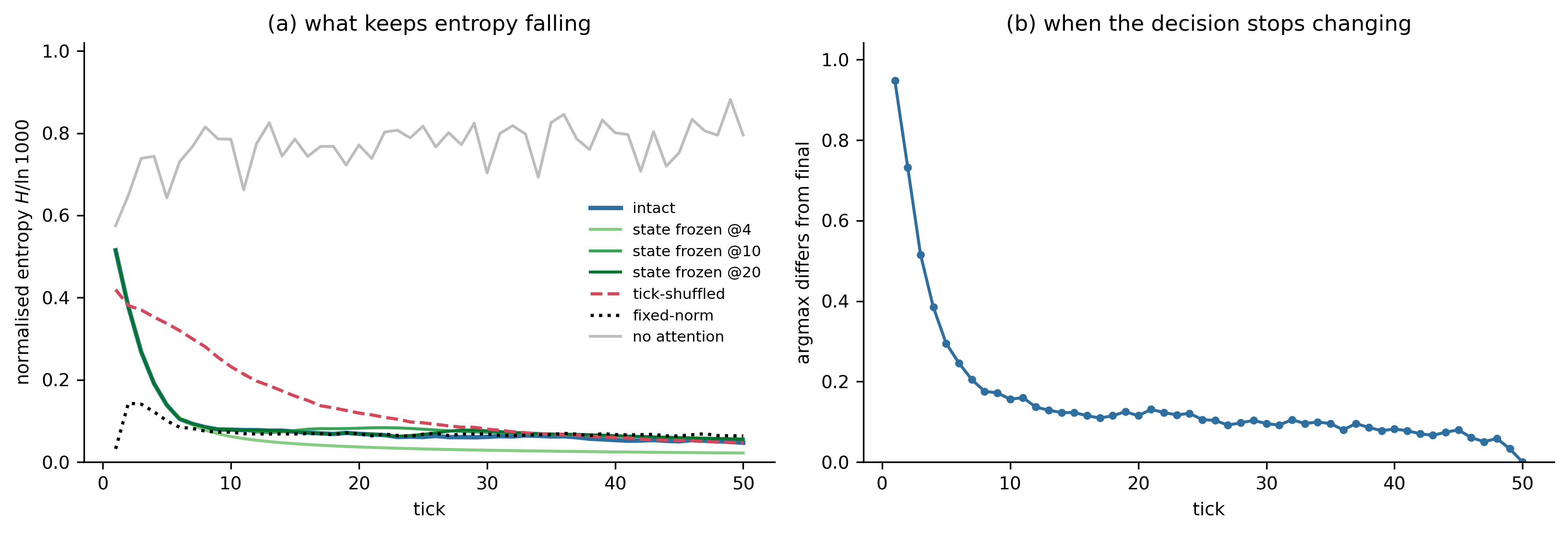}
\caption{\textbf{What sharpens the prediction.} \textbf{(a)}~Normalized entropy over ticks, intact and under each manipulation ($512$ images). \textbf{(b)}~Fraction of images whose predicted class differs from its final value.}
\label{fig:rc3}
\end{figure}

Across the ticks the model grows more certain: normalized predictive entropy ($H/\ln 1000$) falls from $0.52$ to $0.05$ (raw $3.56$ to $0.32$ nats) by the last tick, most of it within the first ten ticks (median $\tdec=39$ of $50$ over $1000$ images; Fig.~\ref{fig:rc3}a). This much is expected---certainty rising over the thought is reported by the CTM's authors and is written into their loss (\S\ref{sec:known}). Our question is what produces it, and the answer is the readout's memory. The decision is read from the synchronization matrix $S^{t}_{\mathrm{out}}$, a decay-weighted running average of pairwise co-activations (Eq.~\ref{eq:st}); in this checkpoint the learned decays are effectively off over the budget, so it is close to a uniform running mean of the state's co-activations. Like a memory it \emph{accumulates and settles}: its per-tick increment $\lVert\Delta S^{t}_{\mathrm{out}}\rVert$ falls and then stays low, even as the state stays live---$\lVert\Delta \zt\rVert$ does not shrink but grows, the traveling activity of \S\ref{sec:waves}. Splitting each tick's change in entropy into the change that would occur if the memory were frozen and only the state moved, and the change caused by the memory updating with the state held fixed, the memory term accounts for all of the drop over $[1,\tdec]$ (it exceeds it: $-0.43$ against a total of $-0.32$ in normalized entropy, $256$ images; 95\% interval for its share $111$--$159\%$), while the state's own motion does not help and slightly raises entropy ($+0.11$); to first order its step is nearly orthogonal to the entropy-reducing direction ($|\!\cos(\Delta\zt,-\gradz)|<0.06$ at every tick).

\paragraph{Causal tests.} On $512$ images (Fig.~\ref{fig:rc3}), three manipulations separate what the memory does from what the state does. \emph{(i) Frozen state.} If the state is held fixed after tick $k$ while the memory keeps accumulating it, entropy keeps falling for $k{=}4$ (to $0.022$ at tick $50$, below the intact $0.047$), yet top-1 at the last tick collapses from $0.73$ to $0.28$; for $k{=}10$ or $20$ entropy stays roughly level and final top-1 falls to $0.02$--$0.03$. Accumulating one repeated state makes the readout confident but wrong: the decision is read from the spread of states along the trajectory, not from any single state. \emph{(ii) Shuffled ticks.} Feeding the same states to the memory in random order changes intermediate entropy but not the final readout ($0.72$ either way), as expected with inert decays: the memory is an order-free average of co-activations. \emph{(iii) Fixed norm.} Rescaling the logits to a common norm removes the descent: normalized entropy stays between $0.03$ and $0.12$ from the first tick on. The entropy fall is therefore largely the growth of the accumulated readout's magnitude, better read as a growing gain than as evidence accumulating. The decision itself settles earlier: the predicted class still differs from its final value for $38\%$ of images at tick $4$, $18\%$ at tick $8$, $12\%$ at tick $16$ and $8\%$ at tick $40$. Without attention, every image receives the same class (here ``football helmet''), since $z^{1}$ is a learned constant and no image information enters.

\section{Smaller models}\label{sec:small}\label{app:small}
Two $D{=}128$ models, one on MNIST and one on CIFAR-10, separate what is intrinsic to CTM inference from
what ImageNet elicits. Both have a dense synchronization matrix: unlike the
ImageNet checkpoint's random pairing, every tracked entry is a pair among the first $16$ neurons, giving
action and output slices of $136$ entries each. The CIFAR model uses a ResNet-9 backbone, which lifts its
accuracy to $0.84$ from $0.64$ with the original backbone and lets us treat it as a competent classifier
rather than a toy.

\paragraph{The curve is universal, the transient amplification is not.} Measured the same way as the ImageNet model,
both have a spectral radius near a tenth at the fixed point ($\rhoJ=0.10$ on MNIST, $0.068$ on CIFAR),
but a largest singular value below one throughout on MNIST ($\sigJ=0.49$, max $0.77$) and in the median
on CIFAR ($0.36$, max $3.06$; about one CIFAR fixed point in ten does stretch)---these maps are Euclidean
contractions, in contrast to the ImageNet map's several-fold stretch (Table~\ref{tab:cross}). They are nonetheless non-normal: already at initialization (normal share $\sum|\lambda|^2/\lVert J\rVert_F^2\approx0.29$, against $0.50$ for a Ginibre matrix), and more so after training ($0.16$ on MNIST, $0.095$ on CIFAR, against $0.020$ on ImageNet). Training thus increases non-normality at every scale, more at larger scale, but only on ImageNet does it reach transient amplification.

\begin{table}[H]
\centering\small
\caption{Cross-model comparison. The attracting curve ($\rhoJ<1$) is universal; transient amplification
($\sigJ\!>\!1$) and query dependence emerge only on ImageNet. ``attn.'' is the entropy of the head-averaged attention weights over image patches, as
a fraction of its maximum $\ln L$ ($1$ = uniform); ``frozen'' and ``shuffled'' are the accuracy changes from freezing the query at its first-tick value and from giving each image another image's query.}
\label{tab:cross}
\begin{tabular}{lcccccc}
\toprule
model & acc. & $\rhoJ$ (med) & $\sigJ$ (med/max) & attn. & frozen & shuffled \\
\midrule
MNIST ($D{=}128$)      & $0.95$ & $0.10$  & $0.49/0.77$ & $0.97$ & $+0\%$ & $+0\%$ \\
CIFAR (ResNet-9)       & $0.84$ & $0.068$ & $0.36/3.06$ & $0.90$ & $+0.3\%$ & $+0.2\%$ \\
ImageNet ($D{=}4096$)  & $0.72$ & $0.15$  & $3.2/74$    & $0.86$ & $-34\%$ & $-7\%$ \\
\bottomrule
\end{tabular}
\end{table}

\paragraph{No attention program, and a briefly aligned descent.} Their attention is close to uniform on MNIST ($H/H_{\max}=0.97$, about $44$ of $49$ patches) and moderately selective on CIFAR ($0.90$, about $42$ of $64$), but on both, freezing or swapping the query moves accuracy by at most a few tenths of a percent, and if anything upwards (frozen/swapped:
$+0.0\%$/$+0.0\%$ on MNIST, $+0.3\%$/$+0.2\%$ on CIFAR), so the query plays no real role: where they attend does not depend on the thought. Only removing
or flattening attention hurts. Entropy falls reliably on both, and there the state's step does line up with the entropy gradient over part of the thought, then turns away from it while entropy keeps falling; the memory term takes over past the crossing. On CIFAR the
alignment is immediate and brief (peak $+0.69$ at tick $1$, crossing at tick $5$), while on MNIST it
builds up and lasts longer (peak $+0.65$ at tick $9$, crossing at tick $17$); on ImageNet the early, aligned phase is absent, and the memory
carries the descent from the start. In all cases the learned forgetting rates confirm that the memory
keeps integrating rather than saturating: the earliest contributions still carry substantial weight at
the decision tick (median retention $0.90$ on MNIST, $0.63$ on CIFAR). Together these place
the small models cleanly: the same skeleton, without the transient amplification or the attention program that
scale brings.

\section{Discussion}
Our results suggest that CTM inference is structured around a common dynamical program shared across images. A moving fixed-point trajectory carries most of the dynamics, while class- and image-specific information appear as offsets from this shared path. The attracting yet strongly non-normal maps we uncover provide a plausible mechanism for transient amplification while maintaining stable dynamics, which would enable task-relevant variations while keeping the state tied to a moving attractor. Non-normal recurrence as a substrate for transient amplification and memory has been studied in neural circuits \citep{ganguli2008memory,murphy2009balanced,goldman2009memory,hennequin2014optimal,bondanelli2020coding}, and built into trained recurrent networks \citep{kerg2019nnrnn}; locating and linearizing around the fixed points of a trained network follows \citet{sussillo2013opening,golub2018fixedpointfinder,maheswaranathan2019reverse}. What differs in the CTM is that the map, and with it the fixed point, is redefined at every tick. 

Our work, however, leaves open the question of what determines the shape of the fixed-point trajectory, and in particular the phase organization giving rise to the CTM's traveling waves. The predictability of attention queries should not be read as evidence that the attention schedule drives the fixed-point curve. Most of the curve's motion persists when the query is frozen, suggesting that the schedule modulates a dynamical process generated largely by the recurrence and its synchronization memory. The fact that attention queries can be predicted makes the CTM more explainable, though it may also limit its flexibility: richer memory dynamics, for example through learned input-dependent decay rates or related fast-weight mechanisms \citep{ba2016fastweights,schlag2021linear}, could enable more image-specific computation. The results reported here are limited to a single checkpoint, a subsample of validation images, and image classification; it remains to be seen whether similar dynamics arise when the CTM is engaged in other tasks.

\bibliographystyle{plainnat}
\bibliography{references}

\newpage
\appendix

\section{Method and robustness}\label{app:fp}

\paragraph{Computing the Jacobian of one thought cycle.}
To study the CTM's neural dynamics we freeze the memory at a chosen tick. With the image tokens $K$, the synchronization memory $S^t$, and the pre-activation history $A^t$ held fixed, one tick becomes an ordinary map from states to states. We collect these frozen quantities into the context \[ \mathcal A^t \equiv (K,S^t,A^t), \] and define \[ z^{t+1}=f(z^t;\mathcal A^t), \] where $\mathcal A^t$ is treated as constant.

We reconstruct this map from the model's own components and verify that a single reconstructed step reproduces the true next state to within numerical error. Iterating the map gives its fixed point; the Jacobian at the fixed point, obtained by automatic differentiation, gives its local behavior. From that Jacobian we read two numbers by power iteration: the spectral radius, which governs whether the map settles, and the largest singular value, which measures how much it can stretch a direction in a single step.

\paragraph{Objects.} At each tick $t$, the frozen map $f(\cdot;\mathcal A^t):\mathbb R^D\to\mathbb R^D$ of Section~\ref{sec:setup} is $C^1$. A \emph{fixed point} $\zstar=\zstar(\A)$ solves
$\zstar=\fmap(\zstar;\A)$; the \emph{Jacobian} there is $J=\partial\fmap/\partial z\big|_{\zstar}$, with
\emph{spectral radius} $\rhoJ=\max_{i}|\lambda_{i}(J)|$ and \emph{operator norm}
$\sigJ=\sqrt{\lambda_{\max}(J^{\top}J)}$. The map is \emph{non-normal} when $J^{\top}J\neq JJ^{\top}$,
which permits $\sigJ>1\ge\rhoJ$: this enables a transient stretching of some directions despite every eigenvalue decaying.

\paragraph{Banach's fixed-point theorem.} If $(X,\|\cdot\|)$ is complete and $g:X\to X$ is a contraction ---
$\|g(u)-g(v)\|\le L\|u-v\|$ for some $L<1$ and all $u,v$ --- then $g$ has a unique fixed point
$x^{\star}$, and the iteration $u_{k+1}=g(u_{k})$ converges to it geometrically,
$\|u_{k}-x^{\star}\|\le L^{k}\|u_{0}-x^{\star}\|$.

\paragraph{How we apply it.} We do \emph{not} have a global Euclidean contraction: $\sigJ>1$, so $\fmap$ can
push nearby points apart in Euclidean norm, and Banach does not apply in $\|\cdot\|_{2}$. Local uniqueness
and attraction follow instead from the spectral radius, through a standard fact (Householder): for any
matrix $J$ and any $\varepsilon>0$ there is a vector norm whose induced operator norm satisfies
$\|J\|_{\star}\le\rhoJ+\varepsilon$. Since we measure $\rhoJ<1$ at every fixed point we find (median $0.15$, max $0.95$), choosing $\varepsilon$ with
$\rhoJ+\varepsilon<1$ yields a norm $\|\cdot\|_{\star}$ in which $J$ is a contraction; because $\fmap$ is
$C^{1}$, it is then a contraction in $\|\cdot\|_{\star}$ on a neighborhood of $\zstar$, and Banach applies
there. The fixed point is therefore locally unique and the iteration converges geometrically in
$\|\cdot\|_{\star}$ --- matching the observation that iterating from the current state reaches residual
$<\!10^{-6}$ at $99.4\%$ of image--tick pairs (see Estimation below). The non-normality ($\sigJ\gg\rhoJ$) is exactly why convergence need not be
monotone in the \emph{Euclidean} norm, and why the state lags the moving fixed point by about one step
even while it is strongly attracted in $\|\cdot\|_{\star}$.

\paragraph{Establishing non-normality.} A stability check based on random directions is therefore misleading: it
averages $\fmap$'s action over directions, reporting something close to the \emph{typical} stretch ---
about $0.07$ here, comfortably below one --- and never samples the few directions stretched most. We
confirm the stretching is genuine, not an artifact of differentiation, by perturbing $\zstar$ along the
single most-stretched direction and measuring the effect by finite differences (over $32$ fixed points, the finite-difference stretch and the autograd operator norm agree to a median ratio of $1.000$; median stretch $2.7$). Stability of an unrolled map like this must be read from
$\rhoJ$ at the fixed point, not from an average stretch.

\paragraph{Estimation.} We estimate $\rhoJ$ and $\sigJ$ two ways that agree in magnitude: power iteration at $\zstar$ (used for Fig.~\ref{fig:spectrum} and the medians $\rhoJ\!\approx\!0.15$, $\sigJ\!\approx\!3.2$, over $3055$ fixed points), and exact eigen- and singular-value decompositions of the fully materialized $J$ (used for the null comparisons of Appendix~\ref{app:schur}; medians over $60$ image--tick pairs, $20$ images at ticks $8$, $24$ and $40$). At the remaining $17$ of the $3072$ image--tick pairs (at ticks $4$, $44$ and $48$) the iteration does not settle within $200$ steps, nor, for the four we examined, within $2000$; no fixed point is found there, and these pairs are excluded from all spectral statistics. Analyses that use fixed points without this check (the lag, the waves, the schedule intervention) report medians, which a fraction this small cannot move appreciably. We quote the non-normality as the median of the \emph{per-fixed-point} ratio $\sigJ/\rhoJ\!\approx\!24$ (95\% interval $23$--$24$) under power iteration; this exceeds the ratio of the marginal medians ($3.2/0.15\!\approx\!21$) because the ratio is right-skewed across ticks. The materialized route gives a somewhat higher median ($\approx\!30$) on its own sample of $60$ fixed points; we use it only for the null comparisons, where measurement and nulls are computed identically.

\paragraph{Other basins.} Banach's theorem establishes local uniqueness only: with $\sigJ>1$ there is no global Euclidean contraction. For $64$ images at ticks $8$, $24$ and $40$, we iterated the frozen map from random states at the typical scale of $\zt$ and at three times it, and from other images' states at the same tick ($1152$ runs of each kind). All $3456$ runs converged, and all reached the fixed point obtained from $\zt$ (median relative distance $2\times10^{-7}$). We found no second basin.

\paragraph{The lag against the contraction rate.} Over $1536$ image--tick pairs ($128$ images), the distance to $\zstar_{t-1}$ shrinks across one step by a median factor $0.11$ (interquartile range $0.06$--$0.22$), against $\rhoJ=0.14$ ($0.08$--$0.33$) at the same fixed points and $0.10$ for the linearized map $\lVert Je\rVert/\lVert e\rVert$; per pair, shrink and $\rhoJ$ are correlated (Spearman $0.66$). The lag is the contraction at work. The fixed point itself moves by a median $37$ per tick, $1.08$ times the pre-step distance; this motion, set mainly by the memory (\S\ref{sec:carrier-main}), is the part of the trajectory the contraction does not imply.

\section{Balanced amplification: the Schur form of the tick map}\label{app:schur}

\paragraph{Pseudospectra and a random-matrix null.} The object behind the transient stretching is the
$\varepsilon$-pseudospectrum of $J$~\citep{trefethen2005spectra}: because $J$ is non-normal, its pseudospectrum reaches well outside the unit disk even though its spectrum sits at radius $\rhoJ\!\approx\!0.15$, which is what allows the one-step Euclidean growth and the lag. Repeated multiplication by a fixed \emph{random} matrix is on its own enough to organize structured neural codes~\citep{hannagan2018random}, so a random matrix is the natural yardstick for an operator applied repeatedly. To ask what produces this non-normality, we compare the measured $\sigJ/\rhoJ$ (of order $30$) against two same-size nulls. A Ginibre matrix (i.i.d.\ Gaussian entries) gives $\sigJ/\rhoJ\!\approx\!2$ by the circular law, so a generic \emph{light-tailed} random matrix is far more normal than $J$.

\paragraph{What produces the non-normality.} An entry-shuffle of $J$---which preserves $J$'s
heavy-tailed entry distribution but destroys its learned arrangement---gives a median $\sigJ/\rhoJ$ of about $13$ over $60$ image--tick pairs, already six times the Gaussian value; the measured $\approx\!30$ adds a further factor of about $2.3$ on top. Both ingredients therefore matter: the heavy-tailed entry \emph{distribution} accounts for most of the gap from a Gaussian matrix, while the specific learned \emph{arrangement} still contributes a substantial additional amount. Either way $J$ is sharply distinguished from a Gaussian matrix of the same size. Part of the non-normality is architectural---an untrained CTM gives $\sigJ/\rhoJ\approx7$---and training multiplies it; training increases it on the small models too, but less (\S\ref{sec:small}). (Medians over $60$ image--tick pairs.)

\begin{figure}[htbp]
\centering
\includegraphics[width=0.6\linewidth]{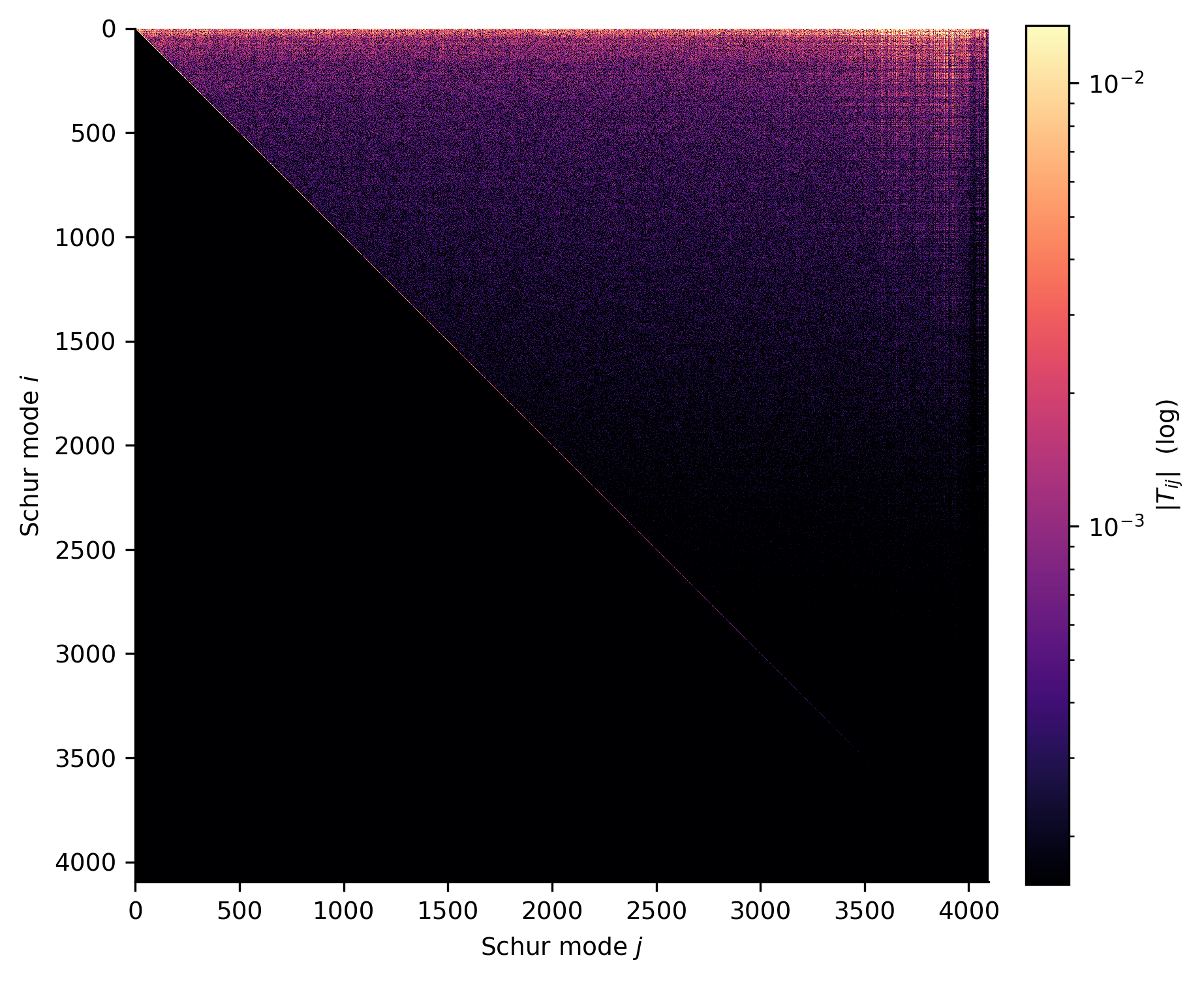}
\caption{\textbf{The tick map is a feed-forward cascade in its Schur basis.} Magnitudes $|T_{ij}|$ (log scale) of the real Schur form of the materialized Jacobian $J=UTU^{*}$ (one held-out image, mid-trajectory tick). The Schur form is upper triangular by construction; what is informative is where the mass sits within it. For a normal operator $T$ would be diagonal, the whole norm carried by the eigenvalues. Here almost none is: the strictly upper-triangular part accounts for a fraction $0.99$ of $\lVert J\rVert_{F}$ (Table~\ref{tab:balamp}), so the operator is nearly all feed-forward coupling between Schur modes rather than modal decay. Magnitudes are shown, and individual mode positions should not be read as structure; see the text.}
\label{fig:schur}
\end{figure}

\paragraph{Balanced amplification.} To probe the \emph{form} of the non-normality we Schur-triangularize the materialized $J=UTU^{*}$ and report three medians (Table~\ref{tab:balamp}): the feed-forward fraction $\lVert\mathrm{triu}(T,1)\rVert_F/\lVert J\rVert_F$ (mass in the strictly upper-triangular, feed-forward part), which equals Henrici's departure from normality $\mathrm{dep}(J)=\sqrt{\lVert J\rVert_F^{2}-\sum_i|\lambda_i|^{2}}/\lVert J\rVert_F$; and the antisymmetric fraction $\lVert(J-J^{\top})/2\rVert_F/\lVert J\rVert_F$. Against the same Ginibre and entry-shuffle nulls, and against the random core (the trained backbone with the recurrence at initialization), $J$ is near-maximally feed-forward ($0.99$ vs $0.71$/$0.81$/$0.93$) with a chance-level rotational part ($0.71$)---the balanced-amplification signature of~\citet{murphy2009balanced}; reading a recurrent operator's hidden feed-forward organization off its Schur form follows~\citet{goldman2009memory}. This structure is visible directly in the materialized operator (Fig.~\ref{fig:schur}).

\paragraph{What the Schur matrix does not show.} Two features of it are conventions rather than findings, and
we read neither. The Schur vectors are each defined only up to sign---$U\mapsto UD$ with $D=\mathrm{diag}(\pm1)$ sends $T_{ij}\mapsto d_{i}d_{j}T_{ij}$---so the sign pattern is gauge-dependent, and we plot $|T_{ij}|$, which is not. The eigenvalue ordering along the diagonal is likewise arbitrary, and reordering it yields a different $U$ and a different $T$ rather than a permutation of the same entries; the placement of individual modes, including the densely coupled leading rows and trailing columns, therefore carries no information. What does not depend on either convention is the diagonal itself---the eigenvalue set in any ordering---and hence the diagonal/off-diagonal split that the feed-forward fraction measures.

\begin{table}[htbp]
\centering\small
\caption{Balanced-amplification descriptors of the tick-map Jacobian $J$ at $\zstar$ (medians over image--tick pairs), against same-size random nulls and the random core (trained backbone, recurrence at initialization; feed-forward fraction only). The Schur feed-forward fraction and Henrici's departure from normality coincide and are reported together. $J$ is a near-maximal feed-forward non-normal operator whose rotational part is at chance.}
\label{tab:balamp}
\begin{tabular}{lcc}
\toprule
 & feed-forward / departure & antisymmetric \\
\midrule
$J$ (trained)        & $0.99$ & $0.71$ \\
Ginibre (i.i.d.)     & $0.71$ & $0.71$ \\
entry-shuffled $J$   & $0.81$ & $0.71$ \\
random core (recurrence at init.) & $0.93$ & --- \\
\bottomrule
\end{tabular}
\end{table}

\section{The attention schedule in detail}\label{sec:carrier}\label{app:carrier}
\begin{figure}[htbp]
\centering
\includegraphics[width=0.95\linewidth]{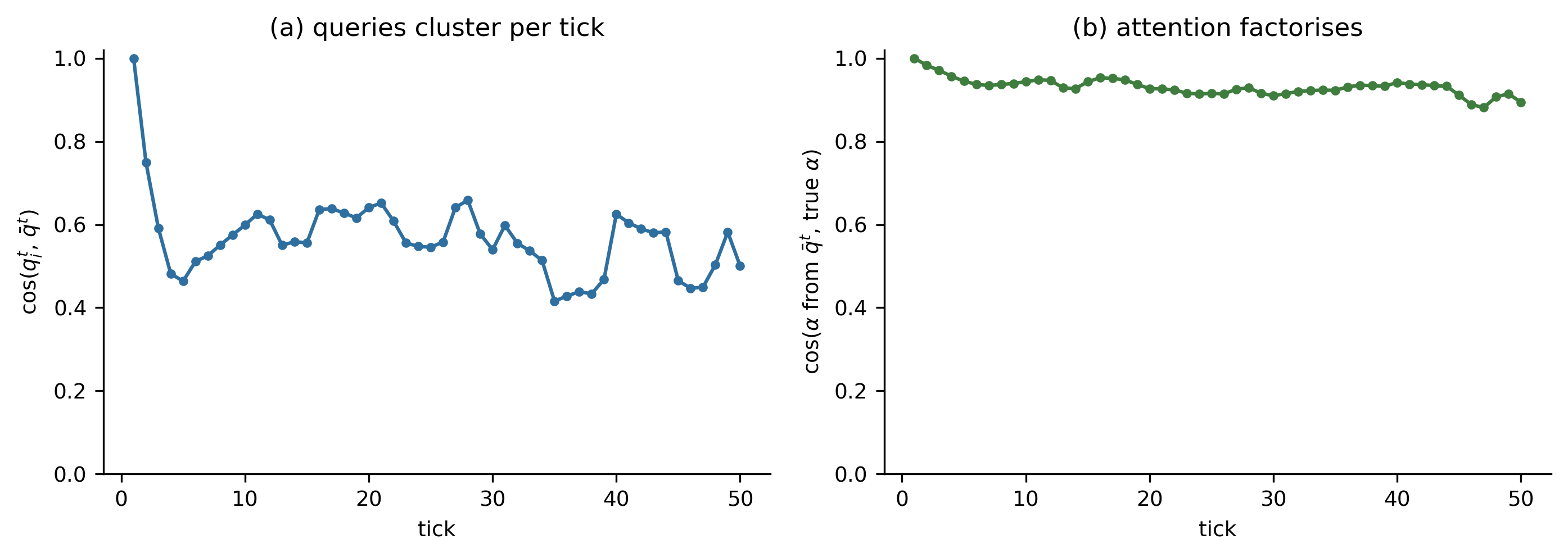}
\caption{\textbf{The shared schedule.} \textbf{Left:} alignment of each image's query with the population-average query at each tick,
$\cos(q^{t}_{i},\bar q^{t})$. \textbf{Right:} the attention computed from the shared,
population-average query matches each image's true attention.}
\label{fig:schedule}
\end{figure}

The attention map comes apart into a shared schedule, which depends only on the tick, and the image's features, which decide where that schedule lands. Discarding each image's own query and substituting the population-average query at the matching tick still recovers the image's true attention map at cosine similarity $0.936$ ($400$ held-out images; Figs.~\ref{fig:scan} and~\ref{fig:schedule}). Although each image sets its own starting query, the \emph{way} the query advances is nearly common to all images, so a simple predictor---the current query plus the shared average increment ($R^{2}=0.98$ held out)---forecasts the next query, and with it the next attention map (cosine $0.996$ versus $0.948$ for assuming attention stays put). Predicting the schedule thus predicts where the model will look.

A note on what this does and does not let one skip at run time. Predicting the next attention map ahead of
time needs only the image's features and the stored schedule; it does not need to run the recurrence.
Predicting the actual class still does, because the decision is read from the state the recurrence
produces. The query's direct effect on entropy, measured as a one-step counterfactual, is modest---it is the routing of attention, compounded over many ticks and captured by the
ablations, that makes the query matter.

\section{Headline numbers}\label{app:controls}
All numbers use the official checkpoint without training; intervals are bootstrap 95\%, and accuracies are computed per image on one common sample of $1000$ images (Table~\ref{tab:numbers}).
\begin{table}[htbp]
\centering\small
\caption{Headline numbers, with the number of images (or image--tick pairs) each rests on and a bootstrap 95\% interval. Accuracies are computed per image on one common sample.}
\label{tab:numbers}
\resizebox{\linewidth}{!}{\begin{tabular}{lccl}
\toprule
quantity & value & 95\% interval & $n$ \\
\midrule
top-1 at $\tdec$ (intact) & $0.723$ & $0.695$--$0.750$ & $1000$ images \\
top-1, frozen / shuffled query & $0.480$ / $0.675$ & $0.450$--$0.511$ / $0.646$--$0.704$ & $1000$ images \\
top-1, uniform / no attention & $0.145$ / $0.001$ & $0.125$--$0.166$ / $0.000$--$0.003$ & $1000$ images \\
$\rhoJ$ at $\zstar$ (median) & $0.147$ & $0.139$--$0.155$ & $3055$ pairs ($256$ images) \\
$\sigJ$ at $\zstar$ (median) & $3.21$ & $3.09$--$3.35$ & $3055$ pairs \\
$\sigJ/\rhoJ$ per fixed point (median) & $23.6$ & $23.0$--$24.3$ & $3055$ pairs \\
distance to $\zstar_{t-1}$ after / before one step & $3.14$ / $33.8$ & $2.82$--$3.44$ / $33.3$--$34.6$ & $1536$ pairs ($128$ images) \\
attention from shared query (cosine) & $0.936$ & $0.934$--$0.940$ & $400$ held-out images \\
next attention predicted ahead / persistence & $0.996$ / $0.948$ & $0.996$--$0.996$ / $0.947$--$0.950$ & $400$ held-out images \\
wave--schedule coupling (CCA, cross-validated) & $0.515$ & null 95th: $0.539$ ($p=0.17$) & $128$ images \\
\bottomrule
\end{tabular}}
\end{table}

\end{document}